\documentclass[journal, twocolumn]{IEEEtran}
\usepackage[english]{babel}
\usepackage[utf8]{inputenc}

\usepackage{xcolor}
\usepackage{graphicx}
\usepackage{subfig}
\usepackage{booktabs}
\usepackage{amsmath, amssymb}
\usepackage{arydshln}
\usepackage{multirow}
\usepackage{icomma}
\usepackage[ruled,vlined]{algorithm2e}

\newcommand{\blue}[1]{\textcolor{black}{#1}}

\begin{document}

\title{Automatic Detection of \emph{Aedes aegypti} Breeding Grounds Based on Deep Networks with Spatio-Temporal Consistency}

\author{Wesley L. Passos{$^{1}$},
Gabriel M. Araujo{$^2$},
Amaro A. de Lima{$^2$},
Sergio L. Netto{$^1$}, and
Eduardo A. B. da Silva{$^1$}.
\thanks{{$^1$}Electrical Engineering Program (PEE/COPPE), Federal University of Rio de Janeiro (UFRJ),
PO Box 68504, Rio de Janeiro, RJ, 21945-970, Brazil.
{$^2$}Federal Center for Technological Education Celso Suckow da Fonseca (CEFET/RJ), Nova Iguaçu, RJ,
26041-271, Brazil.
E-mails:
\{wesley.passos, sergioln, eduardo\}@smt.ufrj.br.
\{gabriel.araujo, amaro.lima\}@cefet-rj.br.}
}

\maketitle

\begin{abstract}
Every year, the \textit{Aedes aegypti} mosquito infects millions of people with diseases such as dengue, zika, chikungunya, and urban yellow fever.
The main form to combat these diseases is to avoid mosquito reproduction by searching for and eliminating the potential mosquito breeding grounds.
In this work, we introduce a comprehensive dataset of aerial videos, acquired with an unmanned aerial vehicle, containing possible mosquito breeding sites.
All frames of the video dataset were manually annotated with bounding boxes identifying all objects of interest.
This dataset was employed to develop an automatic detection system of such objects based on deep convolutional networks. We propose the exploitation of the temporal information contained in the videos by the incorporation, in the object detection pipeline,
of a spatio-temporal consistency module
that can register the detected objects,
minimizing most false-positive and false-negative occurrences.
Also, we experimentally show that using videos is more beneficial than only composing a mosaic using the frames.
Using the ResNet-50-FPN as a backbone,
we achieve F$_1$-scores of 0.65 and 0.77
on the object-level detection of `tires' and `water tanks', respectively, illustrating the system capabilities to properly locate potential mosquito breeding objects.
\end{abstract}


\section{Introduction}
\label{sec:Intro}

The mosquito \emph{Aedes aegypti} is the transmitter of arboviral diseases such as dengue, zika, chikungunya, and urban yellow fever~\cite{ruckert2017}. 
According to the World Health Organization,
from 50 to 100 million dengue infections occur worldwide every year.
In 2020, the Americas alone reported more than 2.3 million cases of dengue,
most of them in Brazil~\cite{web:pahodenguereport}. 
Zika virus infection during pregnancy correlates to microcephaly and other congenital malformation; children in these conditions rarely develop normally~\cite{web:who2018zika}.
It is estimated that there are 200,000 clinical cases of yellow fever, causing 30,000 deaths in the world yearly~\cite{web:pahoyellowfever}.
These diseases also have a strong economical impact.
A survey conducted in 17 countries in the Latin and Central Americas estimates that the cost of dengue epidemics in these countries exceeds US\$ 3 billions annually, US\$ 1.4 billion only in Brazil~\cite{laserna2018economic}. These facts make the arboviruses transmitted by the \emph{Aedes aegypti} one of the leading global health problems.

Except for the yellow fever, there are no vaccines nor specific antiviral drugs for the diseases
transmitted by the \emph{Aedes aegypti}. Thus, the current best form to combat these diseases is
still through the control and elimination of possible mosquito foci~\cite{lambrechts2012vector}.
In fact, the impact of vector reduction on diseases spread has been a subject of study for a long time~\cite{smith1905logical}.
The \emph{Aedes aegypti} reproduces in clean and stagnant water. So, any containers that store water (water tanks, buckets, ornamental fountains, plant dishes, water canisters for animals, tires, and others) are potential breeding grounds. Such objects are very common and can be found everywhere, which can make monitoring and controlling the mosquito, in the absence of proper technical support,  expensive, time-consuming, and inefficient.
Allying the knowledge of an expert with a tool that accelerates the search for potential mosquito foci can be very valuable in such scenario.

\blue{According to current sanitary regulations, health agents must visit properties to search for and eliminate potential mosquito breeding grounds~\cite{web:conass, web:pmvrioliraa}. This approach presents many limitations including temporal or frequency constraints, safety concerns, and costs. Satellite imagery is not considered as a viable alternative due to limited spatial and time resolutions besides its elevated costs~\cite{Grubesic2018UAS}. In that context, unmanned aerial vehicles (UAVs) present several advantages when compared to the previously mentioned approaches. First, not only do UAVs allow for capturing images/videos at higher spatial or time resolution but also at multiple angles, and altitudes. Second, using a programmable UAV increases the auditors’ safety by reducing their exposure to dangerous situations or places. Lastly, although the UAV approach requires a relatively high initial cost for equipment acquisition, it pays off since it is possible to aerial cover a large area with a small team and minimal operating costs~\cite{Grubesic2018UAS}.}
Currently, organizations have been using unmanned aerial vehicles (UAVs) to inspect hard-to-reach sites~\cite{web:pmvitoria2015}. However, usually, an expert has to perform a visual inspection to identify possible objects associated to the mosquito reproduction. This procedure tends to be time-consuming, tiresome, and, consequently, prone to failure.

In this work, we propose the use of images and videos captured by a UAV,
also known as {\it drone},
to support local health agents in locating potential hazardous sites.
We then describe a system to automate the analysis process by applying machine learning and computer vision techniques to aid the specialist in the localization of relevant mosquito foci. 
For this purpose, we present an annotated video database acquired by a UAV flying at different altitudes and following predetermined serpentine-like trajectories.
In addition, using this database, we develop a system capable of detecting these potential productive breeding sites by inputting all video frames to the Faster R-CNN architecture~\cite{Ren2017fasterpami}. We also propose a spatial-temporal consistency module that exploits the correlation of the information present in neighboring video frames by employing the phase correlation technique~\cite{reddy1996phasecorrelation} to estimate the displacements of the detected objects from frame to frame. Using these displacements we perform motion compensation to register the object along the video, eliminating a good portion of false-positive and false-negative detections.

A study~\cite{tun2009reducing} shows that focusing on the most productive breeding sites, such as tires and water tanks, is almost as effective as targeting all possible object classes that are potential mosquito breeding grounds, with the advantage of optimizing time- and cost-related resources. Based on this fact, the focus of this work is on the detection of tires and water tanks.

In order to describe our contributions, this paper is organized as follows: Section~\ref{sec:Revision} describes related works in the subject, whereas Section~\ref{sec:database} describes the newly acquired database. The proposed object-detection system is presented in Section~\ref{sec:system}, where we detail the proposed spatio-temporal consistency module that mitigates most incorrect and missed detections. Section~\ref{sec:methodology} describes the experimental methodology employed to evaluate the proposed system in terms of its performance in detecting the objects of interest, and Section~\ref{sec:results} presents the obtained experimental results. Finally, Section~\ref{sec:conclusion} concludes the paper by emphasizing its main technical achievements.

\section{Related Work}
\label{sec:Revision}

\blue{
A recent review shows that machine learning techniques have gained a lot of attention for mosquito control in urban environments~\cite{Joshi2021review}. Following that trend, the authors of~\cite{bravo2021automatic} also address the detection of potential mosquito breeding grounds using a UAV. They aim to detect water tanks as well as scenarios containing objects that can hold stagnant water. The work in~\cite{Motta2020optimization} proposes a neural network-based model to extract features from the images of mosquitoes to automate the classification among the \textit{Aedes aegypti}, \textit{Aedes albopictus}, and \textit{Culex quinquefasciatus} species. The authors of~\cite{Minakshi2020high} propose an algorithm to detect water bodies that may serve as breeding sites for the main malaria vectors, namely the \textit{Anopheles gambiae} and \textit{Anopheles funestus} species.}

In~\cite{Agarwal2014a}, the authors propose a system to identify potential mosquito breeding sites in geotagged images received from the population. The images are converted into feature vectors using the bag of visual words model through the scale-invariant feature transform (SIFT) descriptor to train a support vector machine (SVM) classifier. According to classification, the system outputs a heat map highlighting the regions with the highest risk of having mosquito habitats. 
The work in~\cite{Mehra2016a} uses an ensemble of naive Bayes classifiers with speeded-up robust features (SURF) extracted from thermal and gray level images to detect stagnant water.
The approaches in~\cite{Agarwal2014a, Mehra2016a} only classify the images as to whether containing or not potential breeding grounds, not providing a precise spatial localization,  basing the analysis on a limited number of object types.

The authors of~\cite{Haddawy2019} use images from Google Street View, Google image search, and Common Objects in Context (COCO) dataset to train and test the Faster R-CNN to detect tires, buckets, potted plants, garbage bins, vases, bowls, and cups. They use the detected objects to compose a dashboard showing the risk areas. In a test set, that system obtained a 0.91 performance in terms of F$_1$-score. 
\blue{
The F$_1$-score is defined as the harmonic mean of precision and recall. Precision is the fraction of correct positive predictions among all predictions, while recall is the fraction of correct positive predictions among all given ground truths. As in precision and recall, the F$_1$-score is limited to the interval [0, 1], where 1 indicates maximum precision and maximum recall, and 0 is obtained if either the precision or the recall is zero~\cite{Paddilla2021}. More details on how to compute precision, recall, and F$_1$-score are given in Section~\ref{sec:eval} (see Eqs.~\eqref{eq:precision}--\eqref{eq:f1}).
}
Despite having promising results, due to the dataset characteristics, the method is only applicable to public places, \textit{e.g.,} streets and avenues. Building on top of the system in~\cite{Haddawy2019}, the authors of~\cite{Prachyabrued2020immersive} designed an immersive visualization tool using a tiled-display wall, aimed at helping researchers and health agents to explore the datasets.

Aerial images can reveal mosquito breeding grounds in abandoned private or difficult to access areas. The papers in~\cite{Passos2018aedes, dias2018mosquitolars} introduced a dataset to enable the design of this type of application. The work in~\cite{dias2018mosquitolars} also proposed a method using optical flow and histograms in the hue, saturation, and value (HSV) color space as feature extractors and random forests to classify tires and stagnant water. Although the obtained results were promising, the dataset used was small and had little variability.

The work in~\cite{Schenkel2020identifying} investigates how a UAV would compare to ground-truth GPS technology for mapping over a small geographical area and identifying relatively small artificial containers, such as bottles, cups, bags, and others, in terms of time, cost, and accuracy. In their study, nine external volunteers collected 678 waypoints data and 214 aerial images from commercial GPS receivers and a UAV, respectively, in an isolated parking lot.
The authors observed that the UAV method was not as efficient as the GPS method for identifying small objects, as the GPS method achieved an F$_1$-score
of 0.86 against 0.30 for the UAV method. 

The works in~\cite{carrasco2019high} and~\cite{haas2019assessing} employed UAVs to detect water bodies that can be breeding sites for \emph{Nyssorhynchus darlingi} or \emph{Cullex Pipiens L.}, respectively. Both works used multispectral and RGB imagery to construct a georeferenced orthomosaic. The former~\cite{carrasco2019high} composed an eight-band orthomosaic and used a random forest algorithm to classify water bodies in the Amazon region with a high chance of having \emph{Nyssorhynchus darlingi} in its aquatic stages. The latter~\cite{haas2019assessing} discriminated between tidal marsh water bodies based on their brightness in the multispectral orthomosaic.
However, the cost and complexity associated with the use of multispectral images limit the applicability of the method.

The recent work in~\cite{bravo2021automatic} addresses the detection of potential mosquito breeding grounds using a UAV by targeting water tanks as well as scenarios containing objects that can hold stagnant water. 
\blue{To conduct their study, the authors of~\cite{bravo2021automatic} compose four datasets,
namely DS1, DS2, DS3, and DS4,
which they make available upon request.}
Datasets DS1 and DS2 contain images with seven types of water tanks acquired from different altitudes
(50~m and 70~m, respectively)
and were used to build and evaluate a YOLOv3 model to detect this object type.
Datasets DS3 and DS4 contain real and simulated scenarios, respectively, acquired with a manually-operated UAV, which can be classified as
critical or normal, depending on the number of
objects of interest present in the scene.

\blue{
In the present work, we extend the dataset initially proposed in~\cite{Passos2018aedes} by adding more videos and many more occurrences of the objects of interest. It also has additional variations regarding recording location, altitude, and object arrangements. The object annotation associated with the videos is dense, that is, it has been performed in all video frames. To the authors’ best knowledge, there is no other database with all of these characteristics that is entirely available to the general public.
}

\blue{
As in previous works, we employ a learning-based object detector algorithm that operates in a frame-level fashion. Specifically, we use the faster region-based CNN (Faster R-CNN)~\cite{Ren2017fasterpami} to detect the objects of interest in each frame. In addition, we enhance the object detection performance by exploiting the temporal redundancy among neighboring video frames. In this process the phase-correlation algorithm~\cite{reddy1996phasecorrelation} is applied along successive frames to align them spatially in a time window. By doing so, we are able to incorporate a spatio-temporal consistency analysis that combines different detections/occurrences of the same object in consecutive frames, thus greatly reducing the number of false and missing detections. In addition, the output of the spatio-temporal consistency model aggregates the results of all frames and gives a detection result that is valid for the whole scene, which obviates the need for further human intervention to generate a detection result that can actually be used in practice.
}

\section{MBG Video Database}
\label{sec:database}

A proper database for detecting and classifying objects in aerial videos should have:
(i) control of the maximum number of recording parameters;
(ii) an expressive number of samples for each object class;
(iii) variability of the background, object position, luminosity, and height;
(iv) no camera distortions; and, last but by no means least, (v) reliable object annotation (position and classification).

In this sense, we present an annotated video database devised for detecting mosquito breeding grounds (MBG) in aerial videos. Drone telemetry (altitude, latitude, longitude, and flying speed) is also available, allowing geolocalization of all objects of interest identified by their corresponding bounding boxes. 
We use the commercial Phantom Vision 4 PRO UAV from DJI Company~\cite{web:djip4prospec} for acquiring the aerial videos using a high-definition camera with passive and active stabilization (dampers and gimbal).

The so-called MBG database has the following technical specifications:
\begin{itemize}
  \item The drone performs a preprogrammed serpentine-like sweep over the entire terrain area autonomously, without any human intervention during video acquisition. Such operating mode
  allows for greater reproducibility and uniformity of recording patterns,
  which becomes independent from the UAV pilot.
  \item Camera auto adjustment is turned off and its
  parameters are set manually, in order to guarantee
  uniformity and total control over the recording process.
  In that manner, camera focus is fixed at infinity and the video scan
  is performed at  $3840\times2160$ or $4096\times2160$ resolutions at 24~frames per second (fps).
  \item Before each flight, a calibration video is recorded using a chessboard pattern to allow compensation for major lens distortions
  using Zhang's method~\cite{Zhang2000}.
  \item The altitude is set to be constant for the whole duration of any video.
  Currently, the database has videos acquired at different altitudes, e.g., $10$, $25$, $40$~m, all of them predefined in the flight plan.
  Small variations in altitude ($\pm 0.5$\,m, according to the manufacturer~\cite{web:djip4prospec}) caused by the limited accuracy of telemetry, wind, etc., are within acceptable ranges.
  Such a height range lends the database a wide applicability: the higher altitudes enable a large area sweep
  during a single video take, while the small altitudes enable the identification of
  small objects such as `tires' and `bottles'.
 \item Speed approximately constant of $15$~km/h is preset in the flying plan, with possible small variations caused by wind, for example.
  \item The dataset includes different types of scenes, such as
  grass, street, buildings, empty lots, and urban areas.
  \item All videos are manually annotated in a frame-by-frame manner
  using the CVAT software~\cite{cvat}.
\end{itemize}

\begin{figure*}[htb!]
    \centering
    \includegraphics[width=0.45\linewidth]{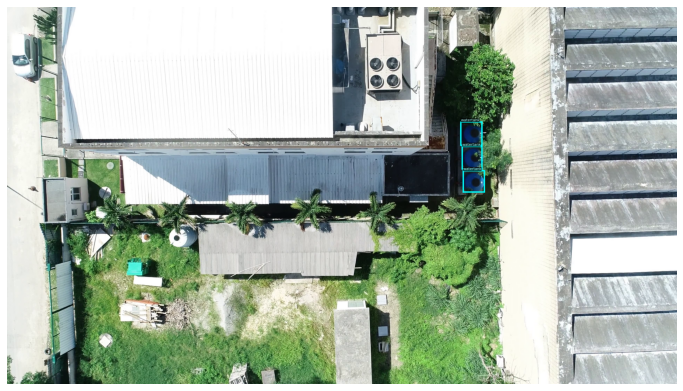}
    \includegraphics[width=0.45\linewidth]{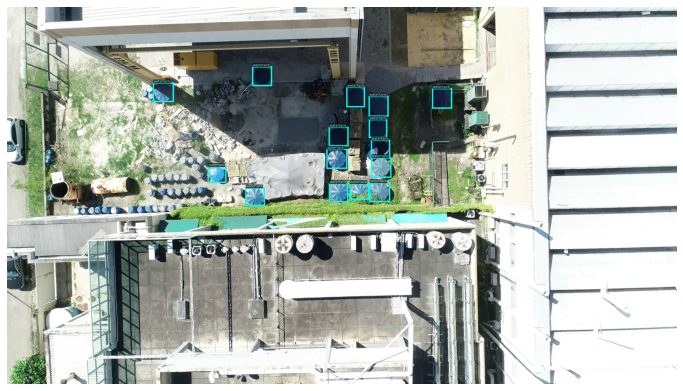}\\ \vspace{0.5mm}
    \includegraphics[width=0.45\linewidth]{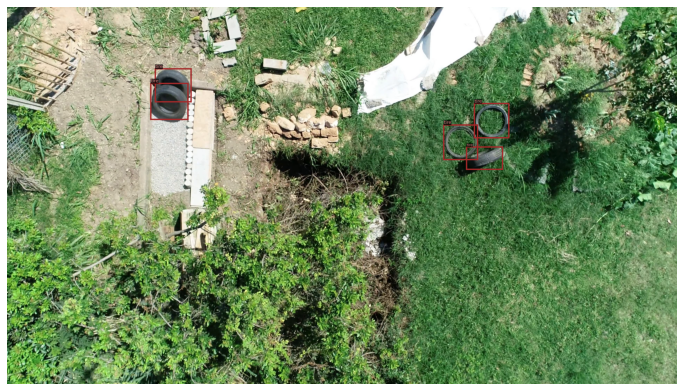}
    \includegraphics[width=0.45\linewidth]{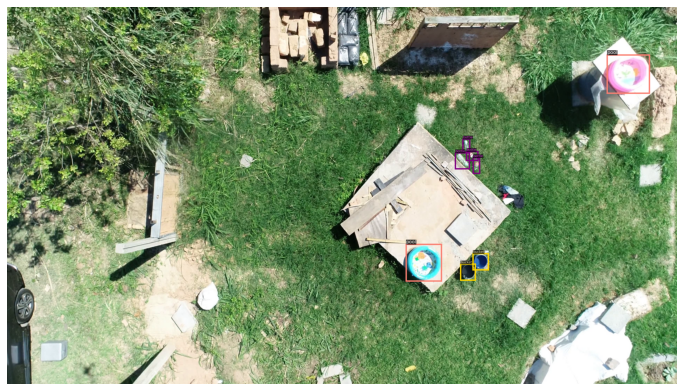}
    \caption{Selected frames of video sequences from the MBG database, showing annotated bounding boxes.}
    \label{fig:dabase}
\end{figure*}

\begin{figure*}[htb!]
\centering
\begin{tabular}{@{}c@{}@{}c@{}@{}c@{}@{}c@{}@{}c@{}@{}c@{}}
    \includegraphics[height=2.2cm]{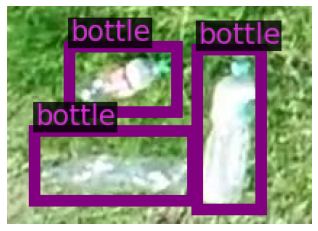} &
    \includegraphics[height=2.2cm]{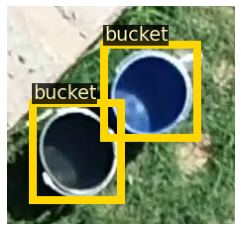} &
    \includegraphics[height=2.2cm]{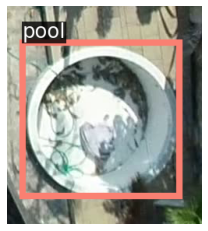} &
    \includegraphics[height=2.2cm]{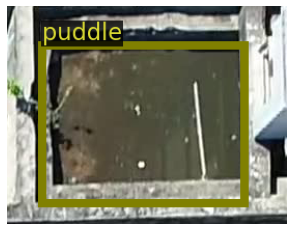} &
    \includegraphics[height=2.2cm]{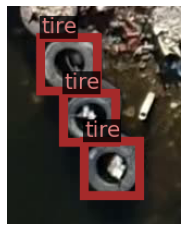} &
    \includegraphics[height=2.2cm]{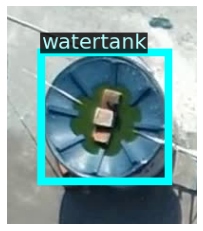}\\
    (a) & (b) & (c) & (d) & (e) & (f) \\
\end{tabular}
    \caption{Examples of objects associated to potential mosquito breeding grounds in the MBG database: 
    (a) `bottle',
    (b) `bucket',
    (c) `pool',
    (d) `puddle',
    (e) `tire', and
    (f) `water tank'.}
    \label{fig:objects}
    
\end{figure*}

The MBG database is publicly available at~\cite{web:mbgdatabase}.
It is presently composed of 13 video sequences acquired by a drone platform recorded in 11 different locations, as detailed in Table~\ref{tab:videos}.
Three of these locations are actual urban areas, while the others had 
the objects of interest manually inserted into the scene in random positions before recording the videos.
Some examples of screenshots of three different MBG videos are shown in
Fig.~\ref{fig:dabase}, whereas Fig.~\ref{fig:objects}
illustrates examples of objects of interest present in the recorded scenes.
\blue{All the videos were recorded in Rio de Janeiro, Brazil.}

\begin{table*}[htb!]
\centering
\caption{Summary of characteristics of each video in the MBG database. The heights are given in meters, the video durations in minutes and the video resolutions are given by the width of the frames in pixels.}
\label{tab:videos}
\begin{tabular}{@{}ccccc@{}}
\toprule
\textbf{Video \#}   & \textbf{Scene composition}      & \textbf{Height {[}m{]}} & \textbf{Duration {[}min{]}} & \textbf{Resolution {[}pixels{]}} \\ \midrule
01            & low grass                       & 10                      & 02:32             & 3840                \\
02            & low grass and street            & 15                      & 00:23             & 3840                \\
03            & low grass, street and buildings & 40                      & 03:44             & 3840                \\
04            & low grass and street            & 15                      & 02:06             & 3840                \\
05            & empty lot                       & 10                      & 00:41             & 3840                \\
06            & empty lot                       & 16                      & 01:15             & 3840                \\
07            & empty lot                       & 20                      & 03:07             & 3840                \\
08            & grass                           & 10                      & 01:37             & 3840                \\
09            & buildings                       & 40                      & 02:41             & 3840                \\
10            & urban zone                      & 40                      & 05:27             & 4096                \\
11            & urban zone                      & 40                      & 05:27             & 4096                \\
12            & urban zone                      & 40                      & 04:33             & 4096                \\
13            & urban zone                      & 40                      & 03:30             & 4096                \\ \bottomrule
\end{tabular}
\end{table*}

The MBG database contains 79,562 frames with a total of 88,281 annotated bounding boxes, comprising several objects such as `bottles', `buckets',
`pools', `puddles', `tires', and `water tanks'.
The large number of annotated bounding boxes enables
one to train object detectors employing deeper networks, which
require large amounts of data for proper parameter adjustment.
Table~\ref{tab:objs_per_video} shows, for each MBG video sequence, the amount of objects of each type and their corresponding number of annotated bounding boxes.

The annotation files are provided in XML format for video tasks as provided by CVAT~\cite{cvat}.
The XML file contains tracks, and each track corresponds to an object which can be presented on multiple frames.
Yet, the objects `bucket,' `poll,' and `water tank' have attributes to indicate object material (plastic or asbestos in `water tank' case), whether it is open or not (`water tank' case), or if it contains water (`pool' and `bucket' cases). 

\blue{
The MBG annotation process was performed by undergraduate students working in groups of three at a time, each student with an average of 10 hours/week. All annotations were later validated in a supervisory round. The entire process took about 10 months for the current MBG version. The CVAT software interpolates the bounding-boxes across non-consecutive frames, greatly speeding up the annotation process. The CVAT also uses previously trained detection models to initialize coarse annotations that can be later refined, what can be used for database expansion with newly acquired videos.
}

\begin{table*}[htb!]
\centering
\caption{Number of annotated object types per MBG video (labeled as `unique'), together with their corresponding number of annotated bounding boxes.}
\label{tab:objs_per_video}
\begin{tabular}{@{}ccccccccccccc@{}}
\toprule
\multirow{2}{*}{\textbf{Video \#}}    & \multicolumn{2}{c}{\textbf{bottle}} & \multicolumn{2}{c}{\textbf{bucket}} & \multicolumn{2}{c}{\textbf{pool}} & \multicolumn{2}{c}{\textbf{puddle}} & \multicolumn{2}{c}{\textbf{tire}} & \multicolumn{2}{c}{\textbf{water tank}} \\ \cmidrule(l){2-13} 
                                   & unique          & bboxes            & unique          & bboxes            & unique         & bboxes           & unique          & bboxes            & unique         & bboxes           & unique             & bboxes              \\ \midrule
01      & 9               & 474               & 2               & 91                & 1              & 53               & 0               & 0                 & 6              & 271              & 0                  & 0                   \\
02      & 7               & 485               & 2               & 135               & 1              & 75               & 0               & 0                 & 6              & 397              & 0                  & 0                   \\
03      & 0               & 0                 & 1               & 161               & 0              & 0                & 0               & 0                 & 0              & 0                & 25                 & 3083                \\
04      & 9               & 283               & 3               & 101               & 3              & 103              & 0               & 0                 & 7              & 339              & 0                  & 0                   \\
05      & 8               & 356               & 4               & 192               & 4              & 171              & 0               & 0                 & 10             & 447              & 0                  & 0                   \\
06      & 1               & 71                & 6               & 405               & 5              & 360              & 0               & 0                 & 6              & 456              & 1                  & 70                  \\
07      & 7               & 605               & 4               & 452               & 3              & 265              & 1               & 86                & 7              & 605              & 0                  & 0                   \\
08      & 11              & 666               & 1               & 214               & 1              & 38               & 0               & 0                 & 6              & 261              & 0                  & 0                   \\
09      & 0               & 0                 & 3               & 281               & 1              & 99               & 2               & 145               & 2              & 192              & 39                 & 2995                \\
10      & 6               & 467               & 26              & 2397              & 5              & 424              & 0               & 0                 & 49             & 4390             & 312                & 25967               \\
11      & 3               & 285               & 10              & 972               & 8              & 839              & 0               & 0                 & 27             & 2625             & 332                & 30855               \\
12      & 0               & 0                 & 2               & 342               & 2              & 339              & 0               & 0                 & 7              & 1278             & 3                  & 383                 \\
13      & 2               & 162               & 2               & 182               & 2              & 185              & 0               & 0                 & 7              & 706              & 0                  & 0                   \\ \midrule
\multicolumn{1}{r}{\textbf{Total}} & \textbf{63}     & \textbf{3854}     & \textbf{66}     & \textbf{5925}     & \textbf{36}    & \textbf{2951}    & \textbf{3}      & \textbf{231}      & \textbf{140}   & \textbf{11967}   & \textbf{712}       & \textbf{63353}      \\ \bottomrule
\end{tabular}
\end{table*}


\section{System Development}
\label{sec:system}

\subsection{Object detection}
Object detection is a classic task in computer vision that consists of both locating and classifying one or more instances of objects in images or videos.
The detector assigns a set of bounding boxes to each image,
containing their coordinates, labels, and confidence scores.

\blue{
Successful object detectors employ convolutional neural networks (CNNs)~\cite{lecun1998procIEEE} in their architectures. The CNNs are inspired by the organization of our visual cortex, where each individual neuron only responds to localized stimuli from our field of vision, and the stimuli of different neurons partially overlap in order to cover the entire field of vision. The CNNs have been successfully applied in many image processing applications due to their ability to automatically extract features from patterns with the least level of pre-processing, which in traditional pattern recognition methods is a handcrafted and laborious process~\cite{lecun1998procIEEE}. CNNs have two main components: convolutional layers and pooling layers.  The convolutional layers consist of a set of learnable filters, that in general are shift invariant, that extract spatial features from images, such as edges, corners etc. These learnable filters reduce the total number of network parameters, besides aiding the CNNs to achieve shift-invariant property, where spatially shifted inputs yield similarly shifted outputs.  The pooling layers, commonly inserted in-between successive convolutional layers, perform spatial sampling operations to produce lower-resolution versions of the convolutional-layer outputs. This also favors shift-invariant representations and reduces the network complexity, thus controlling system overfitting~\cite{goodfellow2016}.}

\begin{figure}[tb!]
    \centering
    \includegraphics[width=\linewidth]{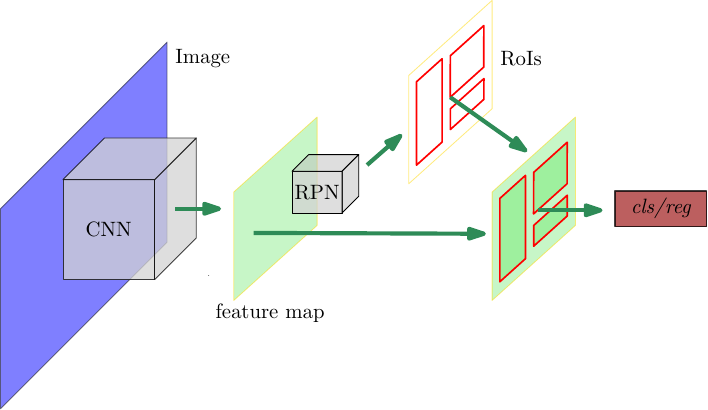}
    \caption{Block diagram for the Faster R-CNN architecture. In this scheme,
    the CNN extracts convolutional feature maps from the input images and the RPN generates potential RoIs. These RoIs are then classified and refined by the classification and regression (cls/reg) module.}
    \label{fig:faster_rcnn}
\end{figure}

The so-called faster region-based convolutional neural network
(Faster R-CNN)~\cite {Ren2017fasterpami}
is a CNN-based meta-architecture that showed excellent results in various
applications and object-detection competitions~\cite{Zhao2019object}.
The Faster R-CNN, as illustrated in Fig.~\ref{fig:faster_rcnn},
is basically composed of a feature extractor, a region proposal generator
(or region proposal network, RPN), together with a classification and regression (cls/reg) module.
In the present work, feature maps are extracted from the input images by using a ResNet-50 convolutional network~\cite{He2016deep} with feature pyramid networks (FPN)~\cite{Lin2017pyramid}.
Based on these maps, the RPN generates potential regions of interest (RoI), which are delivered to the cls/reg module. This module is composed of densely connected layers
that identify the precise RoI boundaries and content classes along with their estimated probabilities. In this work, these probabilities are referred to as confidence scores; commonly, they are used to characterize a detection in the sense that it is considered positive only if its confidence is larger than a threshold $\tau$. 
The detections whose confidence level is less than $\tau$ are discarded.
The confidence threshold $\tau$ is a parameter that is usually set during the system training/validation procedure.

\blue{
In our application, we are concerned about detecting an object in the scene. When using detections in video frames that are regarded as independent of one another, the output is a collection of detections associated with video frames. However, in this case, an object will in general appear in several frames. Besides distorting the results, this requires further interpretation of the results by a human expert. To address this issue, we investigate two approaches. In the first one, a mosaic is composed by selecting frames whose union covers the whole scene with minimum intersection among them. In the second, we use the results of all video frames, but detections of a given object that appears in different frames are aggregated as described in the following section.
}

\subsection{Spatio-temporal tube for object registration}
\label{sec:rastreamento}
When dealing with videos, standard object-detection models do not intrinsically consider the temporal dimension, i.e., each detection of the same object is treated independently across distinct frames.
If, however, one is concerned with whether an object is present in the scene recorded as the whole set of video frames,  as is the case in our application, a post-processing stage is required to combine the bounding boxes of the same object in different frames. 
The works in~\cite{han2016seqnms, kang2018tubelets, belhassen2019seqbbox, sabater2020robust} attempt to combine the detection outputs in several frames to make the object detection results more coherent and stable.
The authors of~\cite{han2016seqnms} attempt to match overlapping detections across frames within the video sequence according to their IOU, without any motion compensation.
The work in~\cite{kang2018tubelets} employs context information from two different object detection models and optical flow to suppress false positives and propagate detections across frames to reduce false negatives.
As it combines the output of two different object detection models, it seems to be computationally complex.
The authors of~\cite{belhassen2019seqbbox} associate overlapping bounding boxes from adjacent pairs of frames to create and re-score object instances and uses them to infer missed detections.
That approach interpolates consecutive frame contents
assuming a linear object movement, which may lead to wrong detections.
The work in~\cite{sabater2020robust} proposes a learning-based
association method for bounding boxes, which does not rely on IOU,
and requires specific data for its training.

In the present work, we consider the presence of an object in the scene as a whole by looking at the entire collection of video frames. We do so in a simple yet effective way, by introducing, in the object detection pipeline, the spatio-temporal tube (STT) concept~\cite{Paddilla2021},
that integrates spatial and temporal localization of the same object. The STT can be viewed as an extension of the 2D spatial bounding boxes present in isolated frames to 3D spatio-temporal bounding boxes defined in a 3D space induced by a set of neighboring video frames.
More precisely, an STT $T_o$ of an object $o$ is the spatio-temporal region defined as the concatenation of the bounding boxes associated with this object in every frame of a video, that is,
\begin{equation}
    T_o = \begin{bmatrix} B_{o,q}\, B_{o,q+1}\, \cdots\, B_{o,q+r} \end{bmatrix}, 
\end{equation}
where $B_{o,k}$ is the bounding box of the object $o$ in frame $k$ of a video that comprises $Q$ consecutive frames indexed by $k = q,\,q+1,\,\ldots,\, (q+Q-1)$. 

The main challenge in the STT approach is how to associate different model outputs with a particular
STT. In this work, we propose a methodology designed taking into account the peculiarities of the MBG database:
\begin{enumerate}[(i)]
\item Displacement between adjacent frames is mainly translational;
\item STT sections in a time interval have the same orientation in 3D space-time, which implies that, although STTs may intersect, they cannot cross each other; and
\item STTs are continuous, which is derived from the fact that a given object cannot appear, disappear, and reappear in consecutive frames.
\end{enumerate}
With these points in mind, an algorithm to compose the STTs
can be implemented according to the following steps:
\begin{enumerate}[I.]
 \item The object spatial displacement between two consecutive frames is estimated by a global motion estimation algorithm such as the phase-correlation~\cite{reddy1996phasecorrelation}. Global motion estimation algorithms are good choices in this case because one can safely assume that, for the objects of interest (see Figure~\ref{fig:objects}),  parallax effects are negligible at the drone heights in which the videos of the MGB database have been acquired (see Table~\ref{tab:videos}). 
 \item 
The computed displacement is employed to estimate the spatial position of the objects 
across different frames and, as a result, one can associate each bounding box to
a particular STT.
 \item 
Having aligned the frames spatially,
we compute the pairwise intersection of union (IOU)
of the detected bounding boxes in two consecutive frames.
We consider that the bounding boxes from two consecutive frames belong to the same STT $T_n$ and, consequently, to the same object, if the IOU between these two bounding boxes is maximum and greater than 0.5.
More precisely, given frames $q$ and $q+1$ aligned by global motion estimation, bounding boxes $B_{n, q}$ and $B_{m, q+1}$,
in frames $q$ and $q+1$, respectively, are associated
to the same STT $T_n$ if the following condition holds:
\begin{eqnarray}
\text{IOU}(B_{m, q+1}, B_{n, q})\geq \text{IOU}(B_{o, q+1}, B_{n, q}) \geq 0.5, ~ \forall o. \!\!
\end{eqnarray}
If this condition does not hold,
we may have either a missed (false negative) or an incorrect (false positive) detection,
which can be discarded altogether.
\end{enumerate}

\begin{figure}[tb!]
    \centering
    \includegraphics[width=0.8\linewidth]{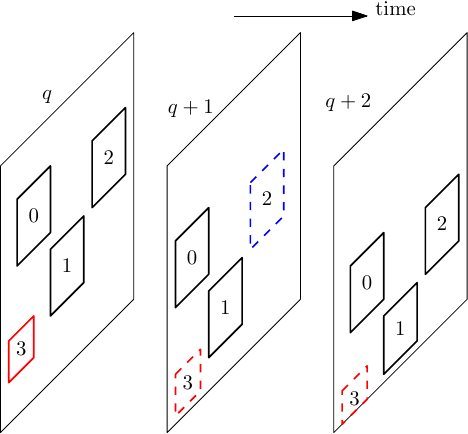}
    \caption{Example of bounding box associations to constitute spatio-temporal tubes (STTs). 
     Solid black lines are detections (model output) over these frames and dashed lines are projected detections from a previous frame.
    The collection of detections and/or estimates having the same object identity form an STT.}
    \label{fig:post_processing}
\end{figure}

\begin{figure}[tb!]
    \centering
    \includegraphics[width=\linewidth]{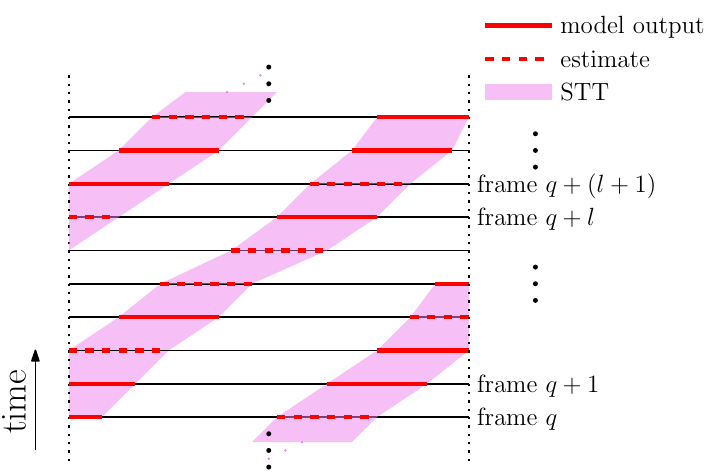}
    \caption{STT representation (pink spatio-temporal regions) across consecutive video frames
    (indicated by black solid lines).
    Solid red lines are detections (model output) over these frames
    and dashed red lines are projected detections from a previous frame.
    The collection of detections and/or estimates having the same object identity
    form a given STT.}
    \label{fig:post_processing2}
\end{figure}

To illustrate the procedure, let us consider the example depicted in Fig.~\ref{fig:post_processing},
where the rectangles with the same number are detections of the same object in different frames (bounding boxes with  \mbox{$\max(\text{IOU}) \geq 0.5$} among aligned frames).
In this sense, object 2 characterizes a false negative (dashed-blue bounding box) in frame $q+1$, while object 3 characterizes a false positive (solid-red bounding box) in frame $q$.
The idea is to identify the detections associated with the same object along the whole video, as depicted in Fig.~\ref{fig:post_processing2},
where the object spatio-temporal localization in the video is represented by the pink path.

Let $l_i$ be the number of detections output by the model within ${\rm STT}_{i}$, and let $m_i$ be the number of frames comprising ${\rm STT}_{i}$, that is, the number of frames contained in the interval between the first frame and the last frame in which the object that is associated with ${\rm STT}_{i}$ is detected.
In order to evaluate if a given ${\rm STT}_{i}$ is associated to
an actual object of interest, we test $(l_i/m_i)$ against a consistency threshold $\mu$. If $(l_i/m_i) \geq \mu$, then ${\rm STT}_{i}$ is equivalent to a detected object. 
The  STT consistency threshold $\mu$ is a parameter to be tuned in the system training/validation process.

\section{Experimental Methodology}
\label{sec:methodology}

\subsection{Performance evaluation}
\label{sec:eval}
To evaluate the network detections it is necessary to establish an IOU threshold $t$, which will determine each detection as correct (true positive, TP) or incorrect (false positive, FP).
One metric widely used in the context of object detection is the
average precision with IOU $\geq 0.5 $ (AP$_{50}$)~\cite{Paddilla2021}, that is, TPs and FPs are defined with $t=0.5$.
The AP metric aggregates the compromise between TPs and FPs for all confidence thresholds $\tau$. In the application associated with this work, that is the detection of mosquitoes breeding grounds, it is also useful to report particular values of
TP, FP, and missed (false negative, FN) detections. In addition, to aggregate the TP, FP, and FN information, we also report precision (Pr),
recall (Rc), and F$_1$-score. Since one considers as positives only those detections whose confidence is larger than a confidence threshold $\tau$, these can be expressed as~\cite{Paddilla2021}:
\begin{eqnarray}
    \text{Pr($\tau$)} &\!\!\!=\!\!\!& \frac{  \displaystyle \sum_{n=1}^S\text{TP}_n(\tau)}{  \displaystyle \sum_{n=1}^S\text{TP}_n(\tau) + \sum_{n=1}^{N-S}\text{FP}_n(\tau)}, 
    \label{eq:precision} \\
    \text{Rc($\tau$)} &\!\!\!=\!\!\!& \frac{  \displaystyle \sum_{n=1}^S\text{TP}_n(\tau)}{  \displaystyle \sum_{n=1}^S\text{TP}_n(\tau) + \sum_{n=1}^{G-S}\text{FN}_n(\tau)}, 
    \label{eq:recall} \\[0.2cm]
    {\rm F}_1(\tau) &\!\!\!=\!\!\!& 2 \frac{\text{Pr}(\tau)  \text{Rc}(\tau)}{\text{Pr}(\tau) + \text{Rc}(\tau)},
    \label{eq:f1}
\end{eqnarray}
where $G$ is the number of ground-truths in the dataset and $N$ is the number detections output by the model, of which $S$ are correct ($S \leq G$).

The aforementioned metrics evaluate the system performance at a bounding-box level. However, as discussed in Subsection~\ref{sec:rastreamento}, when dealing with videos, one may be mostly interested in evaluating the system performance at the object level, using STTs. In this case, one can use the STT-AP metric, proposed in~\cite{Paddilla2021}, which is computed in a way equivalent to the one in which the AP metric is computed, with the difference that the bounding boxes are replaced by STTs.
In that manner, an object is considered a TP if the STT-IOU is equal or greater than a chosen threshold $t_{\rm T}$, which was set to $t_{\rm T}=0.5$ in this work, yielding the metric that we refer to as STT-AP$_{50}$.

Likewise the case of the AP metric, in the STT-AP all the model predictions are ranked for each class according to the predicted confidence level $\tau$ (from the highest to the lowest), irrespective of their correctness.
The confidence level $\tau_{\rm T}$ assumed for an STT is the average confidence of the bounding boxes corresponding to each of its constituent frames.
The all-point interpolation~\cite{Paddilla2021} may be then performed in order to compute the final STT-AP value.
We also report the TP, FP and FN as well as the Pr, Rc and F$_1$ at STT level, which can be computed using Eqs.~\eqref{eq:precision},~\eqref{eq:recall}, and~\eqref{eq:f1} by using both $\tau$ and $\mu$ as thresholds.
\blue{
It is important to mention that the focus of the following experiments is on detecting tires and water tanks, as these objects are considered the most productive containers for the \textit{Aedes aegypti} species~\cite{tun2009reducing}.
}

\subsection{Training the Faster R-CNN \blue{using all the video frames for frame-level metrics}}\label{sec:FasterRCNNframes}
For the sake of development of the proposed mosquito breeding grounds detection system,
the MBG database was initially split into training and test sets.
The five videos exclusively assigned to the test set
(01, 02, 05, 09, and 13) are only used for final model evaluation.
Targeting model development and hyper-parameter tuning,
a leave-one-out cross-validation scheme~\cite{bishop2006book}
was employed. In this procedure, the training set was partitioned in eight folds,
as shown in Table~\ref{tab:loocv},
where seven videos were used for the actual training
and the remaining one was used for validation.

\begin{table}[th!]
\centering
\caption{Training and validation sets used in the experiments with the MBG dataset.
Videos 01, 02, 05, 09, and 13 were exclusively assigned to the test set.}
\label{tab:loocv}
\begin{tabular}{@{}ccc@{}}
\toprule
\multirow{2}{*}{\textbf{Fold}} & \multicolumn{2}{c}{\textbf{Video} \#}                                                                      \\ \cmidrule(l){2-3} 
                & Training                      & Validation \\ \midrule
1               & 04, 06, 07, 08, 10, 11, 12    & 03        \\
2               & 03, 06, 07, 08, 10, 11, 12    & 04        \\
3               & 03, 04, 07, 08, 10, 11, 12    & 06        \\
4               & 03, 04, 06, 08, 10, 11, 12    & 07        \\
5               & 03, 04, 06, 07, 10, 11, 12    & 08        \\
6               & 03, 04, 06, 07, 08, 11, 12    & 10        \\
7               & 03, 04, 06, 07, 08, 10, 12    & 11        \\
8               & 03, 04, 06, 07, 08, 10, 11    & 12        \\ \bottomrule
\end{tabular}
\end{table}

All models are initialized with the public available weights~\cite{wu2019detectron2} trained on the COCO database and are further trained in the MBG database
with a learning rate of $0.002$.


In the case the system was tuned with the frame-level metrics, having the eight best models in each training and validation fold, 
the confidence interval $\tau$ of the Faster R-CNN was tuned by exhaustive search in the set from $\tau=0.1$ to $\tau=0.9$ in increments of $0.1$ with the added confidence intervals $\tau=0.05$ and $\tau=0.99$, according to the 
best F$_1$ value averaged over all validation sets.
The best confidence interval was found to be $\tau=0.99$ for both the `tire' and `water tank' cases.

For the final model, once the confidence threshold $\tau$ was set,  
the system was retrained with all eight videos in the original training set.
To mitigate overfitting, this final training process is interrupted at iteration $I$,
which is the median value of the iteration indexes $I_i$, for $i=1,2,\ldots,8$
that correspond to the minimum loss-function value in each validation run.

\blue{\subsection{Training the Faster R-CNN using the Frame Mosaic}
As previously mentioned, one may be concerned with the presence of an object in the scene. To this end, we compose what we call a frame mosaic. To build it, we manually sample the frames from each video, ensuring that the selected frames cover the entire scene and at the same time each object of interest appears only once in the whole frame collection. Having built the frame mosaic, the models are trained as described in Section~\ref{sec:FasterRCNNframes} and the confidence threshold $\tau$ of the Faster R-CNN is tuned. The confidence threshold values found were $\tau = 0.8$ and $\tau = 0.5$ for `tire’ and `water tank’ classes, respectively.}

\subsection{Tuning the STT hyper-parameter} 

The procedure described in Section~\ref{sec:rastreamento}
was applied to generate the STTs,
aggregating the object-related temporal and spatial information present in the videos. In order to tune the model according to STT-based metrics, hyperparameters were searched for, namely the confidence threshold $\tau$ of the Faster R-CNN and the consistency threshold $\mu$ of the STT (see Subsection~\ref{sec:rastreamento}). After grid search, the confidence threshold was set to $\tau=0.99$, and the model consistency threshold was set to $\mu = 0.4$ 
for both `tire' and `water tank' object classes
by maximizing the average F$_1$-score for the STTs across all eight validation folds.

\section{Experimental Results and Discussion}
\label{sec:results}

\subsection{Frame-level object detection}
The frame-level performance of each of the eight Faster R-CNN trained models
was evaluated in the test set, as given in
Tables~\ref{tab:results_tire} and~\ref{tab:results_watertank}  for the
`tire' and `water tank' object classes, respectively,
presenting satisfactory AP$_{50}$ results in both cases.

\begin{table}[th]
\caption{Frame-level, bounding-box based results on the test set for object class `tire', considering an IOU threshold $t=0.5$ and $\tau=0.99$.}
\label{tab:results_tire}
\resizebox{\linewidth}{!}{%
\begin{tabular}{@{}cccccccc@{}}
\toprule
\textbf{Fold}     & \textbf{AP$_{50}$} & \textbf{TP} & \textbf{FP} & \textbf{FN} & \textbf{Pr} & \textbf{Rc} & \textbf{F$_{1}$} \\ \midrule
1                 & 61.06         & 53          & 9           & 27          & 0.85        & 0.66        & 0.75      \\
2                 & 58.69         & 50          & 6           & 30          & 0.89        & 0.63        & 0.74      \\
3                 & 59.52         & 51          & 5           & 29          & 0.91        & 0.64        & 0.75      \\
4                 & 50.19         & 47          & 9           & 33          & 0.84        & 0.59        & 0.69      \\
5                 & 55.84         & 47          & 5           & 33          & 0.90        & 0.59        & 0.71      \\
6                 & 64.79         & 55          & 7           & 25          & 0.89        & 0.69        & 0.77      \\
7                 & 58.29         & 49          & 3           & 31          & 0.94        & 0.61        & 0.74      \\
8                 & 51.36         & 48          & 16          & 32          & 0.75        & 0.60        & 0.67     \\
\midrule
\textbf{average} & 57.47          & 50.00       & 7.50        & 30.00       & 0.87        & 0.63        & 0.73      \\
\textbf{std}     & 4.87           & 2.88        & 4.00        & 2.88        & 0.06        & 0.04        & 0.04      \\ \midrule
\textbf{final}   & 58.57          & 51          & 5           & 29          & 0.91        & 0.64        & 0.75    \\\bottomrule
\end{tabular}%
}
\end{table}
\begin{table}[!h]
\caption{Bounding-box based results on the test set for object class `water tank', considering an IOU threshold $t=0.5$ and $\tau=0.99$.}
\label{tab:results_watertank}
\resizebox{\linewidth}{!}{%
\begin{tabular}{@{}cccccccc@{}}
\toprule
\textbf{Fold}     & \textbf{AP$_{50}$} & \textbf{TP} & \textbf{FP} & \textbf{FN} & \textbf{Pr} & \textbf{Rc} & \textbf{F$_{1}$} \\ \midrule
1 & 64.12 & 102 &  48 &  25 & 0.68 & 0.80 & 0.74  \\
2 & 66.85 & 103 &  39 &  24 & 0.73 & 0.81 & 0.77  \\
3 & 70.16 & 108 &  50 &  19 & 0.68 & 0.85 & 0.76  \\
4 & 65.60 & 108 &  54 &  19 & 0.67 & 0.85 & 0.75  \\
5 & 65.15 & 105 &  44 &  22 & 0.70 & 0.83 & 0.76  \\
6 & 58.03 & 101 &  63 &  26 & 0.62 & 0.80 & 0.69  \\
7 & 67.33 & 109 &  62 &  18 & 0.64 & 0.86 & 0.73  \\
8 & 70.16 & 109 &  50 &  18 & 0.69 & 0.86 & 0.76  \\
\midrule
\textbf{average} & 65.93 & 105.63  & 51.25   & 21.38    & 0.67  & 0.83  & 0.74      \\
\textbf{std}     & 3.87  & 3.29    & 8.26    & 3.29     & 0.04  & 0.03  & 0.02      \\ \midrule
\textbf{final}       & 65.46  & 105  & 44 & 22 & 0.70 & 0.83 & 0.76               \\ \bottomrule
\end{tabular}%
}~
\end{table}

Overall, the models were able to identify correctly the majority of the `tires' in the MBG
dataset presenting a very low number of FPs.
The 16 FP occurrences in fold 8, a figure much larger than the ones of the other folds,
are probably due to the validation video being a bit darker than the training ones.
This hampered the network ability to distinguish between round shadows and tires.
In addition, most of the FN figures correspond to tires
that were placed at particularly challenging positions,
as discussed in Subsection~\ref{subsec:stt_level_results}.
In the case of `water tanks',
the models show satisfactory results in terms of F$_1$,
which is a good compromise between Pr and Rc.

The last rows in Tables~\ref{tab:results_tire} and~\ref{tab:results_watertank} show the results for the final model obtained by training the Faster R-CNN using all eight available training videos.
As one may observe, in this last row all resulting performance metrics are within the mean~$\pm$~std interval of the set of folds, indicating a stable and robust training procedure.
Also, the final model performed better in the test set in terms of F$_1$,
indicating that the network benefits from more data.

In the case of frame-level metrics, as the same object repeatedly appears along several frames,
it contributes more than once for the TP, FP, and FN scores.
This distortion is mitigated with the STT-level analysis,
whose results are discussed in the following subsection.

\blue{\subsection{Object detection using the Frame Mosaic} \label{subsec:mosaic_results}
The performance of each of the eight Faster R-CNN trained models was evaluated in the frame mosaics built from the videos in the test set, as given in Tables~\ref{tab:mosaic_tire} and~\ref{tab:mosaic_watertank} for the `tire' and `water tank' object classes, respectively. One may notice the large number of false positives which implies a low precision. This may be explained by the low variability of the objects in the training sets allied to the fact that each object appears only in one frame, which causes the model to have a single chance to detect each object in the scene. In case the image of a mosaic frame is degraded through drone movement, which is common to occur~\cite{Wu2021survey}, the object may be missed.}

\begin{table}[b!]
\caption{\blue{Mosaic results on the test set for object class `tire', considering an $t=0.5$ and $\tau=0.8$.}}
\label{tab:mosaic_tire}
\resizebox{\linewidth}{!}{%
\blue{
\begin{tabular}{@{}cccccccc@{}}
\toprule
\textbf{Fold}     & \textbf{AP$_{50}$} & \textbf{TP} & \textbf{FP} & \textbf{FN} & \textbf{Pr} & \textbf{Rc} & \textbf{F$_{1}$} \\ \midrule
1 & 56.04 & 22 & 40 & 9  & 0.35  &0.71 & 0.47 \\
2 & 54.57 & 21 & 32 & 10 & 0.40 & 0.68 &  0.50 \\
3 & 43.87 & 22 & 31 & 9  & 0.41  &0.71 & 0.52 \\ 
4 & 48.70 & 21 & 23 & 10 & 0.48 & 0.68 &  0.56 \\ 
5 & 43.01 & 15 & 18 & 16 & 0.45 & 0.48 &  0.47 \\ 
6 & 61.44 & 20 & 17 & 11 & 0.54 & 0.64 &  0.59 \\ 
7 & 53.96 & 21 & 30 & 10 & 0.41 & 0.68 &  0.51 \\ 
8 & 56.17 & 21 & 21 & 10 & 0.50 & 0.68 &  0.58 \\
\midrule
\textbf{average}& 52.22 & 20.37 & 26.50 & 10.62 & 0.44 & 0.66 & 0.52 \\
\textbf{std}    & 6.44  & 2.26  & 8.02  & 2.26  & 0.06 & 0.07 & 0.04 \\ \midrule
\textbf{final}  & 57.35 & 21    & 19    & 10    & 0.52  & 0.68 & 0.59 \\ \bottomrule
\end{tabular}%
}~}
\end{table}

\begin{table}[b!]
\caption{\blue{Mosaic results on the test set for object class `water tank', considering an $t=0.5$ and $\tau=0.5$.}}
\label{tab:mosaic_watertank}
\resizebox{\linewidth}{!}{%
\blue{
\begin{tabular}{@{}cccccccc@{}}
\toprule
\textbf{Fold}     & \textbf{AP$_{50}$} & \textbf{TP} & \textbf{FP} & \textbf{FN} & \textbf{Pr} & \textbf{Rc} & \textbf{F$_{1}$} \\ \midrule
1 & 56.51 & 30 & 22 & 9 & 0.58 & 0.77 & 0.66 \\ 
2 & 63.82 & 35 & 28 & 4 & 0.56 & 0.90 & 0.69 \\ 
3 & 58.93 & 34 & 26 & 5 & 0.57 & 0.87 & 0.69 \\ 
4 & 59.29 & 33 & 25 & 6 & 0.57 & 0.85 & 0.68 \\ 
5 & 68.09 & 36 & 45 & 3 & 0.44 & 0.92 & 0.60 \\
6 & 55.28 & 33 & 41 & 6 & 0.45 & 0.85 & 0.58 \\
7 & 58.92 & 35 & 49 & 4 & 0.42 & 0.90 & 0.57 \\ 
8 & 61.72 & 34 & 30 & 5 & 0.53 & 0.87 & 0.66 \\
\midrule
\textbf{average}& 60.32 & 33.75 & 33.25 & 5.25 & 0.51 & 0.87 & 0.64 \\
\textbf{std}    & 4.13 & 1.83 & 10.22 & 1.83 & 0.07 & 0.05 & 0.05  \\ \midrule
\textbf{final}  & 51.32 & 31 & 28 & 8 & 0.52 & 0.79 & 0.63 \\ \bottomrule
\end{tabular}%
}~}
\end{table}

\subsection{STT-level object detection} \label{subsec:stt_level_results}

In this section, bounding boxes that 
correspond to the same object along consecutive frames are counted as a single tube,
as detailed in Subsection~\ref{sec:rastreamento},
and all performance metrics are determined at the STT level.
Tables~\ref{tab:res_tube_tire} and~\ref{tab:res_tube_watertank} show the results for the object detection considering the video as a whole, using STTs.
From these tables, one readily notices a drastic reduction in all FP values
due to the temporal consistency imposed by the STT concept.

\begin{table}[b!]
\caption{STT based test results on the test set for object class `tire', considering an STT-IOU threshold $t_{\rm T}=0.5$ and consistency threshold $\mu=0.4$.}
\label{tab:res_tube_tire}
\resizebox{\linewidth}{!}{%
\begin{tabular}{@{}cccccccc@{}}
\toprule
\textbf{Fold} & \textbf{STT-AP$_{50}$}& \textbf{TP} & \textbf{FP} & \textbf{FN} & \textbf{Pr} & \textbf{Rc} & \textbf{F$_{1}$}  \\ \midrule
1             & 58.21        & 18          & 1           & 13          & 0.95        & 0.58        & 0.72                 \\
2             & 61.24        & 19          & 1           & 12          & 0.95        & 0.61        & 0.75                 \\
3             & 54.46        & 17          & 0           & 14          & 1.00        & 0.55        & 0.71                 \\
4             & 51.49        & 16          & 3           & 15          & 0.84        & 0.52        & 0.64                 \\
5             & 51.16        & 16          & 2           & 15          & 0.89        & 0.52        & 0.65                 \\
6             & 60.59        & 19          & 3           & 12          & 0.86        & 0.61        & 0.72                 \\
7             & 54.29        & 17          & 1           & 14          & 0.94        & 0.55        & 0.69                 \\
8             & 53.97        & 17          & 2           & 14          & 0.89        & 0.55        & 0.68                 \\\midrule
\textbf{average} & 55.67     & 17.38       & 1.63        & 13.63       & 0.92        & 0.56        & 0.69                 \\
\textbf{std}     & 3.89      & 1.19        & 1.06        & 1.19        & 0.05        & 0.04        & 0.04                 \\ \midrule
\textbf{final}   & 51.48     & 16          & 2           & 15          & 0.89        & 0.52        & 0.65 \\\bottomrule
\end{tabular}%
}
\end{table}
\begin{table}[b!]
\caption{STT based test results on the test set for object class `water tank', considering an STT-IOU threshold $t_{\rm T}=0.5$ and consistency threshold $\mu=0.4$.}
\label{tab:res_tube_watertank}
\resizebox{\linewidth}{!}{%
\begin{tabular}{@{}cccccccc@{}}
\toprule
\textbf{Fold} & \textbf{STT-AP$_{50}$}& \textbf{TP} & \textbf{FP} & \textbf{FN} & \textbf{Pr} & \textbf{Rc} & \textbf{F$_{1}$}  \\ \midrule
1                &  71.61         & 33          &  18         &   6         &  0.65       &  0.85       &  0.73  \\
2                &  71.81         & 32          &  17         &   7         &  0.65       &  0.82       &  0.73  \\
3                &  67.04         & 34          &  20         &   5         &  0.63       &  0.87       &  0.73  \\
4                &  61.83         & 34          &  21         &   5         &  0.62       &  0.87       &  0.72  \\
5                &  74.08         & 33          &  15         &   6         &  0.69       &  0.85       &  0.76  \\
6                &  69.82         & 35          &  22         &   4         &  0.61       &  0.90       &  0.73  \\
7                &  62.47         & 34          &  27         &   5         &  0.56       &  0.87       &  0.68  \\
8                &  65.95         & 35          &  21         &   4         &  0.63       &  0.90       &  0.74  \\\midrule
\textbf{average} & 68.08          & 33.75       & 20.13       & 5.25        & 0.63        & 0.87        & 0.73   \\
\textbf{std}     & 4.50           & 1.04        & 3.64        & 1.04        & 0.04        & 0.03        & 0.02   \\ \midrule
\textbf{final}   & 71.33          & 35          & 17          & 4           & 0.67        & 0.90        & 0.77 \\ \bottomrule
\end{tabular}%
}
\end{table}

\begin{figure*}[th!]
    \centering
    \begin{tabular}{@{}c@{}@{}c@{}@{}c@{}@{}c@{}@{}c@{}}
    \includegraphics[height=3.5cm]{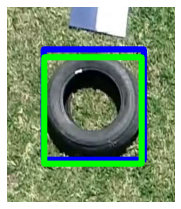} &
    \includegraphics[height=3.5cm]{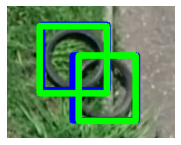} &
         \includegraphics[height=3.5cm]{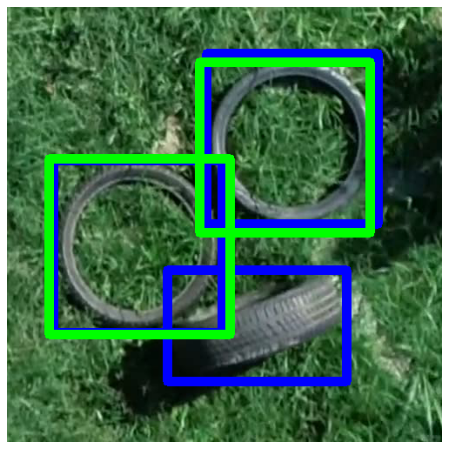} &
         \includegraphics[height=3.5cm]{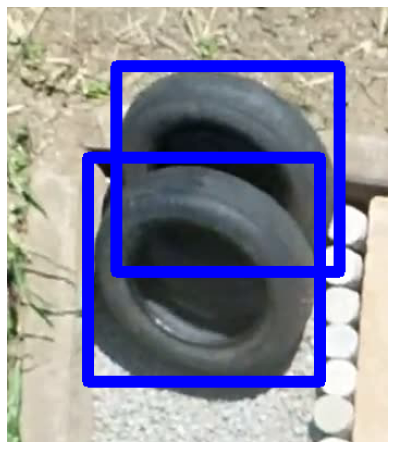} &
         \includegraphics[height=3.5cm]{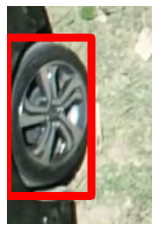}\\
    \end{tabular}
    \caption{Example results of `tires' detection, with some manual annotation (blue), correct detection (TP, in green), and false detection (FP, in red).
    Note that the blue boxes not overlaid by a green box are false negatives (FN).}
    \label{fig:res_tires}
\end{figure*}

For the `tire' object class, one may note the high TP scores and low FP values,
leading to a high system precision.
Some examples of correct `tire' detections are depicted in Fig.~\ref{fig:res_tires},
where one observes how the system is able to detect successfully
tires of very different thicknesses.
Most FN cases for the `tire' class are associated with the target
objects in the vertical position or stacked, thus causing partial occlusion,
as illustrated in Fig.~\ref{fig:res_tires}.
Such cases are underrepresented in the training set.
An interesting case of FP is also shown in Fig.~\ref{fig:res_tires}
in which the trained model detected the tire that is part of the car.

\begin{figure*}
    \centering
    \begin{tabular}{@{}c@{}@{}c@{}@{}c@{}@{}c@{}@{}c@{}}
         \includegraphics[height=3.5cm]{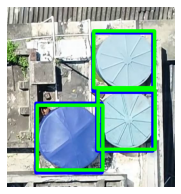} &
         \includegraphics[height=3.5cm]{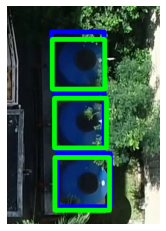} &
         \includegraphics[height=3.5cm, width=3cm]{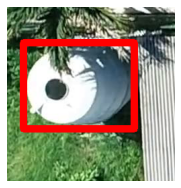} &
         \includegraphics[height=3.5cm]{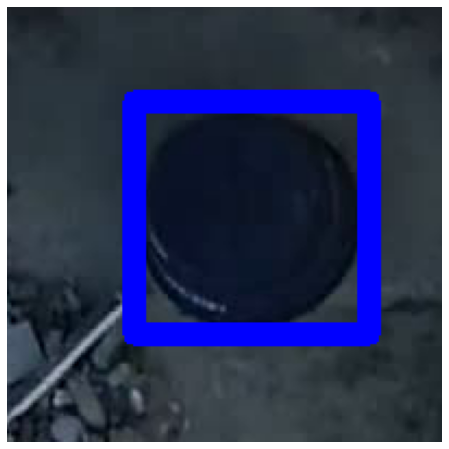}&
         \includegraphics[height=3.5cm]{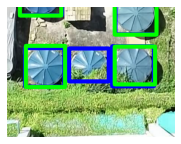}\\
    \end{tabular}
    \caption{Example results of `water tank' detection, with some manual annotation (blue), correct detection (TP, in green), and false detection (FP, in red).
    Note that the blue boxes not overlaid by a green box are false negatives (FN)}
\label{fig:res_watertank}
\end{figure*}

In the `water tank' class,
the system is capable of retrieving 90\% of the objects present in the test set,
including several cases of different standards and colors,
as illustrated in Fig.~\ref{fig:res_watertank}.
Most FP occurrences in this class are due to objects that, even to the human eye,
can be easily mistaken for `water tanks',
as also depicted in Fig.~\ref{fig:res_watertank}.
In addition, most FN cases are related to the object
being positioned in the shadow, occluded, or even upside-down.
All these detection problems may be mitigated by possible database extension to incorporate
more instances of these critical situations.

\blue{
The drone movement may cause some aerial images to be fuzzy or noisy, which may lead to image degradation~\cite{Wu2021survey}. Using videos, one may increase the chances of the object to be detected, as this object may appear at different points of view. As one may observe, the STT approach allows to improve performance using a temporal context. Comparing the results obtained obtained with a frame mosaic with the ones obtained using an STT, we verify that there has been an improvement in the performance by 10.17\% and  22.22\% for `tire’ and `water tank’ classes, respectively, in terms of F$_1$-score. Also, one may notice that the STT approach leads to a more stable system, with a lower F$_1$-score standard-deviation across folds.}

\blue{In Brazil, the annual survey of \textit{Aedes aegypti} infestation has been mandatory since 2017. The health ministry established that all Brazilian cities must annually send two surveys, the ``Survey of Sample Index'' (LIA) and ``Rapid Survey of \textit{Aedes aegypti} Infestation Index'' (LIRAa), to the federal government~\cite{web:conass}. The local agents perform the surveys by taking the possible breeding grounds into account~\cite{web:pmvrioliraa}. Then, the local health managers use these reports to decide the allocation of the agents. The problem is that there are many breeding grounds such as water tanks or tires in hard-reaching areas. These objects usually are not in the report because they are above constructions or inside a private property that the ground agents cannot see or enter. An aerial image-based report can complement the others and help the local manager to make better decisions when allocating the human resource. 
} 

\blue{On the other hand, aerial mapping is faster than a ground-based survey and requires a smaller team. In addition, as shown in the results, the proposed method performs automatic detection of potential mosquito breeding grounds, which can potentially speed up the analysis process performed by the health agents, increasing its overall efficiency and effectiveness in finding and eliminating such breeding grounds. In the long run, such productivity improvement may lead to reduced insect proliferation and disease propagation.}

\blue{
In this work, we balance the importance of both FPs and FNs simultaneously by optimizing the system according to the F$_1$-score. This is so because the errors have an impact on resource allocation. An FP can induce the manager to send local agents to an area with no mosquito infestation. On the other hand, an FN can induce the manager to neglect an area with potential to cause an outbreak. It is important to point out that it is possible to adjust the system to be more or less sensitive to FPs in detriment of FNs or vice versa. In the present application, for instance, FNs are considered more significant than FPs, that is, the miss  of a potential breeding site tends to be far more damaging than raising a false alarm.
}

\blue{
\subsection{System strengths and limitations}
The system proposed in this work is able to automatically detect potential mosquito breeding sites. Specifically, we focus on two types of objects, namely water tanks and tires, which are considered the highest productive containers for the \textit{Aedes aegypti} species~\cite{tun2009reducing}. Although focused on these object classes, this work may be extended to include object groups if so desired or required. The UAV-based analysis allows for a wider area coverage and its pre-planned flying pattern enables different operators to reproduce the same recording procedure in different locations even with different drone models. Achieved results indicate that the proposed system can successfully detect objects of interest in an automatic manner, thus improving health agents’ efficiency in locating and subsequently eliminating mosquito breeding grounds.
}

\blue{
The system also presents some limitations. As pointed out above, in its current version, there is some drop in the detection efficiency when analyzing darker images. The system also tends to fail in distinguishing a round shadow from a tire. The former issue may be addressed in future system developments by increasing the number of darker images in the training set. The latter issue may be addressed using for example a post-processing step based on color space analysis. Another noticeable concern occurs when an object is occluded by another object of interest, such as two superimposed tires. The current system only detects one of the objects in these situations; although it does not constitute a major practical issue.
}

\section{Conclusion}
\label{sec:conclusion}

In this work, the automatic detection of {\em Aedes aegypti} mosquitoes foci using videos acquired with an autonomous UAV was addressed. A database of aerial images that contains several objects commonly associated with potential mosquito breeding grounds was introduced.
A complete system for detecting objects of interest was proposed.
\blue{
Focus was given to water tanks and tires, as these two object types are considered the most critical ones for the \textit{Aedes aegypti} reproduction. If required, other object types can be readily addressed following the same system development methodology described in this paper. The proposed system works} in two stages: 
first, object detection is performed frame-by-frame
using the Faster R-CNN architecture;
later, a subsequent step registers successive detections and imposes
spatio-temporal consistency to the detection process
which is performed at the object level.
\blue{This last step has shown to be more robust when compared to the results of the frame mosaic.}
Using the ResNet-50-FPN as a backbone,
it was possible to obtain an F$_1$-score of 0.65 and 0.77
for the `tire' and `water tank' classes, respectively, which are considered the most
critical objects for the \textit{Aedes aegypti} wide reproduction.
As can be inferred from our results, there is good evidence that the system can be improved with more training data. However, as discussed in this paper, acquiring and performing through bounding box annotation in such data is not an easy task.
As a future direction, one may use active or self-learning to help in the annotation process of future acquired videos.
Artificial data augmentation using generative adversarial neural networks could also lead to improved results.

\section*{Acknowledgment}
This study was financed in part by the Coordenação de Aperfeiçoamento de Pessoal de Nível Superior – Brasil (CAPES) – Finance Code 001; Fundação Carlos Chagas Filho de Amparo à Pesquisa do Estado do Rio de Janeiro (FAPERJ); Conselho Nacional de Desenvolvimento Científico e Tecnológico (CNPq); and Google Latin America Research Awards (LARA), 2019--2020 and 2020--2021.
The authors would like to express their gratitude to the students from CEFET/RJ \textit{campus} Nova Iguaçu whose work was fundamental to the database annotation process.

\bibliographystyle{IEEEtran}
\bibliography{IEEEabrv,references}

\end{document}